\documentclass[conference,letterpaper]{IEEEtran}
\IEEEoverridecommandlockouts
\usepackage[font=footnotesize]{caption}
\usepackage{subcaption}
\usepackage[font=footnotesize]{caption}
\usepackage[table]{xcolor}

\usepackage{amsmath} 
\usepackage{amssymb}
\usepackage{breqn}
\usepackage{newunicodechar}
\newunicodechar{α}{\alpha}
\newunicodechar{γ}{\gamma}

\usepackage{algorithm}
\usepackage{algorithmicx}
\usepackage{algpseudocode}

\usepackage{cite}

\usepackage{titlesec}
\usepackage{booktabs}
\usepackage{graphicx}
\graphicspath{{figures/}}
\usepackage[export]{adjustbox}

\usepackage{tikz}
\usepackage{array}
\usepackage{pbox} % used in table of phase 3

\algnewcommand{\EndComment}[1]{{\color{RawSienna}\hfill \(\triangleright\) #1}}
\algnewcommand{\LineComment}[1]{{\Statex \color{RawSienna} \(\triangleright\) #1}}

\usepackage{balance}

\usepackage{comment}
\usepackage{xcolor}

\definecolor{R}{RGB}{0,130,0}
\title{\LARGE \bf
Robust Small Methane Plume Segmentation in Satellite Imagery 
}

    \author{Khai Duc Minh Tran$^{1}$, Hoa Van Nguyen$^{1}$, Aimuni Binti Muhammad Rawi$^{2}$, Hareeshrao Athinarayanarao$^{2}$, Ba-Ngu Vo$^{1}$% <-this % stops a space 
    \thanks{This research is supported by the ARC Linkage Project LP200301507.}% <-this % stops a space
    \thanks{$^1$School of EECMS, Curtin University, Bentley WA 6102, Australia.
            {\tt\footnotesize \{minh.tran2; hoa.v.nguyen; ba-ngu.vo\}@curtin.edu.au}.}%
    \thanks{$^2$Latconnect60, 1 William St, Perth WA 6000, Australia.
            {\tt\footnotesize \{aimuni; hareeshrao\}@latconnect60.com}.}
}

\begin{document}

\maketitle
\thispagestyle{empty}
\pagestyle{empty}

%%%%%%%%%%%%%%%%%%%%%%%%%%%%%%%%%%%%%%%%%%%%%%%%%%%%%%%%%%%%%%%%%%%%%%%%%%%%%%%%
\begin{abstract}
This paper tackles the challenging problem of detecting methane plumes, a potent greenhouse gas, using Sentinel-2 imagery. This contributes to the mitigation of rapid climate change. We propose a novel deep learning solution based on U-Net with a ResNet34 encoder, integrating dual spectral enhancement techniques (Varon ratio and Sanchez regression) to optimise input features for heightened sensitivity. A key achievement is the ability to detect small plumes down to 400 m² (i.e., for a single pixel at 20 m resolution), surpassing traditional methods limited to larger plumes. 
Experiments show our approach achieves a 78.39\% F1-score on the validation set, demonstrating superior performance in sensitivity and precision over existing remote sensing techniques for automated methane monitoring, especially for small plumes. 
\end{abstract}

%%%%%%%%%%%%%%%%%%%%%%%%%%%%%%%%%%%%%%%%%%%%%%%%%%%%%%%%%%%%%%%%%%%%%%%%%%%%%%%%
\section{Introduction}  

Methane (CH$_4$) is the second most influential anthropogenic greenhouse gas after carbon dioxide, responsible for approximately a quarter of observed global warming since pre-industrial times \cite{vaughan2023ch4net,radman2023s2metnet,lauvaux2022global}. Due to its relatively short atmospheric lifetime of about 9-12 years, achieving rapid reductions in methane emissions can deliver near-term climate benefits, making methane mitigation a critical strategy for slowing warming in the coming decades  \cite{pandey2023daily}. The urgency of methane control is underscored by the accelerating rise in atmospheric methane concentrations, driven primarily by human activities such as fossil fuel production, waste management, and agriculture \cite{alvarez2018assessment}.

Anthropogenic methane emissions exhibit a highly skewed, heavy-tailed distribution, where a small subset of large, anomalous point sources (``super-emitters") accounts for a disproportionate share of total emissions \cite{lauvaux2022global, pandey2023daily}. While episodic ultra-emission events from oil and gas infrastructure offer cost-effective mitigation targets \cite{lauvaux2022global, pandey2023daily, alvarez2018assessment}, a narrow focus on these super-emitter events overlooks the significant, cumulative impact of smaller, persistent, and under-reported emissions from other sectors. For instance, recent studies indicate substantially higher methane releases from inefficient gas flaring than previously assumed \cite{plant2022inefficient}. 
Furthermore, satellite observations combined with machine learning are revealing the widespread impact of the agricultural sector, linking emissions to specific dairy farm practices and enabling new mitigation pathways \cite{bi2024mapping}. 
Capturing this emission's ``long tail'' is essential for accurate global methane accounting and for formulating comprehensive climate mitigation strategies that address the full spectrum of emission sources.

Satellite remote sensing is instrumental in monitoring anthropogenic methane emissions \cite{SnchezGarca2021MappingMP, varon2020high, ruuvzivcka2023semantic}. Global monitoring relies on instruments like the TROPOspheric Monitoring Instrument (TROPOMI) aboard Sentinel-5P, which offers daily global coverage but at a coarse spatial resolution of $5.5 \times 7$ km\textsuperscript{2} \cite{varon2022continuous}. Conversely, high-resolution airborne hyperspectral imagers such as the Airborne Visible/Infrared Imaging Spectrometer (AVIRIS-NG) enable detailed, targeted plume analysis \cite{jongaramrungruang2022methanet}. This presents a trade-off between the temporal revisit rate achieved by spaceborne sensors and the fine-scale spatial detail offered by airborne platforms. The Sentinel-2 constellation, with its unique combination of global coverage, a revisit rate of $2-5$ days, and $20\times20$ m\textsuperscript{2} resolution in key Short-Wave Infrared (SWIR) bands, is strategically positioned to bridge this observational gap between frequent monitoring and high-resolution detail~\cite{vaughan2023ch4net}. Although not specifically designed for direct gas detection, Sentinel-2's SWIR bands exhibit sensitivity to methane absorption features, rendering them capable of detecting larger plume signatures \cite{Groshenry2022DetectingMP}.

Despite this potential, existing physics-based retrieval methods for Sentinel-2 often produce high false-positive rates, especially over heterogeneous surfaces, and require extensive manual validation \cite{ruuvzivcka2023semantic}. Detecting smaller, more persistent leaks is particularly challenging due to low signal-to-noise ratios \cite{vaughan2023ch4net, radman2023s2metnet, pandey2023daily}. 
In this work, we address these challenges by introducing a deep learning approach for methane plume segmentation in Sentinel-2 imagery, with a focus on detecting small-scale emissions. Our primary contributions are:
\begin{itemize}
\item[i)] A novel feature engineering approach, specifically designed to optimise the input features of satellite imagery, that combines two spectral enhancement techniques to improve the detection of methane signals.
\item[ii)] An evaluation of deep learning architectures demonstrating that a U-Net with a ResNet34 encoder, when combined with our novel feature engineering approach, overcomes the limitations of existing methods to detect single-pixel methane plumes in Sentinel-2 images.
\end{itemize}

\section{Related Work}

Methane monitoring via satellite has advanced from physics-based retrievals, using radiative transfer and methane's spectral absorption models, to complementary deep learning models.

\textbf{Multi-Sensor Methane Detection via Tiered Satellite Systems.} Emerging solutions focused on tiered observation systems that combine data from multiple satellites to balance spatial resolution, temporal coverage, and detection sensitivity. For example, integrated systems leverage Sentinel-5P for daily global methane monitoring, Sentinel-2 for high-resolution (20 m) detection, and Sentinel-3 as an intermediate sensor \cite{pandey2023daily}. The use of multi-band, multi-pass (MBMP) retrieval algorithms with the SWIR bands of Sentinel-2 has proven capable of quantifying large leaks. Varon et al. \cite{varon2020high} further demonstrated high-frequency monitoring of anomalous methane point sources with Sentinel-2, showing that operational satellites can reliably detect significant emissions by applying robust ratioing techniques. At much finer scales, very high-resolution sensors like WorldView-3 (3.7 m) have shown promise in mapping methane from specific infrastructure \cite{SnchezGarca2021MappingMP}, while airborne imaging spectrometers such as AVIRIS-NG serve as critical tools for algorithm development and validation at even higher resolutions \cite{ruuvzivcka2023semantic}.

Despite their successes, traditional physics-based approaches face several challenges. Methane absorption signals in multispectral data are relatively weak and can be confounded by surface reflectance heterogeneity and varying illumination geometry \cite{ruuvzivcka2023semantic}. As a result, simple difference or ratio methods often trigger false positives over spectrally complex landscapes. Even sophisticated regression techniques may require expert intervention to distinguish real plumes from surface artifacts \cite{SnchezGarca2021MappingMP}. Cloud edges and thin cirrus also pose a problem, frequently creating spurious ``plume-like" signals in SWIR images. Consequently, these methods typically require manual post-processing and cross-verification \cite{ruuvzivcka2023semantic}, which reduces scalability and increases response times. While effective for large plumes, they often fail to reliably detect the ``long tail" of smaller or more subtle emission events.

\textbf{Rise of Deep Learning Models.} 
Recently, deep learning (DL) has become a powerful alternative to classical methane detection methods. In hyperspectral imaging, Ruzicka et al. \cite{ruuvzivcka2023semantic} introduced the STARCOP dataset, achieving significantly higher F1-scores than physics-based models by learning complex spectral patterns. For multispectral data, the S2MetNet project established a benchmark dataset, showing that DL models outperform traditional methods in quantifying methane emissions from Sentinel-2 \cite{radman2023s2metnet}. Similarly, Vaughan et al. \cite{vaughan2023ch4net} proposed CH4Net, a U-Net-based model trained on extensive annotated Sentinel-2 imagery, setting a new standard for methane detection by effectively distinguishing real plumes from false positives.

Beyond single-model approaches, researchers are exploring techniques like data-efficient deep transfer learning to adapt models to new geographic domains with high accuracy \cite{Zhao2025ADD} and ensemble methods that combine multiple models or multi-resolution observations to increase detection reliability \cite{Ayasse2024ProbabilityOD}. While these DL approaches represent a major leap forward, accurately detecting small, persistent leaks near the detection limit of current sensors and ensuring model generalisation across diverse conditions remains an active research area. Our work builds directly on this foundation by focusing on small-plume detection through refined feature engineering %and by providing a new, comprehensive real-world dataset 
to facilitate further progress.

\section{Dataset and Methods}
Our primary objective is to develop an automated and scalable system for the detection and segmentation of methane plumes from Sentinel-2 imagery, with a specific focus on identifying small plumes that are often missed by existing methods. This requires overcoming three core challenges: (i) dataset generation incorporating feature extraction techniques to support the detection of small methane plumes in multispectral data, (ii) creating a high-quality, large-scale annotated dataset of real-world plumes, and (iii) selecting and optimising a DL architecture sensitive enough for this task.

\subsection{Feature Enhancement}
A high-quality dataset is critical for training robust deep learning models. To this end, we constructed a dataset comprising real plume events, methane-free background scenes, and augmented variations. All data were sourced from Sentinel-2 imagery of an oil and gas facility in Australia, a site with well-documented CH$_4$ releases in 2023.

\textbf{Feature Enhancement:} 
The methane absorption characteristic in its short-wave infrared (SWIR) bands extracted from Sentinel-2 multispectral data is subtle and often obscured by background surface reflectance \cite{ruuvzivcka2023semantic}. To overcome this, we transformed the raw satellite radiances into a multi-channel image that maximises the contrast between methane-plume and background pixels. This transformation is achieved by calculating two spectral ratios on a per-pixel basis, each generating a spatial map of ratio values - a single-channel image that highlights different aspects of the methane signal. These individual channels are then combined into a multi-channel representation. Let $R_{11}$ and $R_{12}$ be the top-of-atmosphere (TOA) radiances for Sentinel-2's SWIR bands 11 and 12, respectively. Band 12 (2190 nm) is sensitive to methane absorption, while nearby Band 11 (1610 nm) is largely unaffected, making it an ideal reference.

\textbf{Varon Ratio $V(\cdot,\cdot)$:} 
    This ratio amplifies the methane signal by normalising the strongly methane-absorbing band $R_{12}$ against the weakly absorbing band $R_{11}$ \cite{varon2020high}. It is calculated for each pixel as:
    \begin{equation}
    V(R_{12},R_{11})=\dfrac{cR_{12}-R_{11}}{R_{11}}.
    \end{equation}
    Here, the scaling factor $c$, computed via a least-squares fit of $R_{12}$ against $R_{11}$, accounts for the combined effects of instrumental response and uniform surface spectral albedo. In pixels affected by a methane plume, atmospheric methane absorbs incoming solar radiation, suppressing $R_{12}$ radiance, while $R_{11}$ remains unaffected. The Varon ratio leverages this spectral contrast to highlight methane-rich pixels. The resulting output is a single-channel image in which where methane plumes become clearly visible as spatially coherent regions with notably different pixel intensities compared to the background.
    
\textbf{S\'{a}nchez Ratio $S(\cdot)$:} 
    While effective, the Varon ratio can still be sensitive to certain surface types (water bodies) having naturally low $R_{12}$ to $R_{11}$ ratios, leading to false positives. The S\'{a}nchez ratio, inspired by S\'{a}nchez-Garc\'{i}a et al. \cite{SnchezGarca2021MappingMP}, uses a multi-linear regression model trained on plume-free imagery to estimate the expected radiance, $\hat{R}_{12}$, from other non-absorbing bands (Band 11 and other visible/SWIR bands), representing the radiance expected for that surface in the absence of methane. 
    The S\'{a}nchez ratio is given by~\cite{ruuvzivcka2023semantic}:

    \begin{equation}
      \label{eq:sanchez-swir}
      S(R_{12}) = V(R_{12},\hat{R}_{12}).
      \end{equation}
      By replacing the single reference band $R_{11}$ with a more robust, surface-adaptive prediction $\hat{R}_{12}$, this approach effectively isolates the plume signal from background spectral variations. This yields a single-channel image where, similar to the standard Varon ratio, methane plumes become easily distinguishable from normal background variation.

By stacking these two single-channel images, we formed a two-channel image where the Varon ratio (V) highlights methane signals, and the S\'{a}nchez ratio (S) reduces noise by filtering out false positives from background variability. Since many DL models require three-channel input, we duplicate V to form a pseudo-RGB image [V, S, V]. 
This setup reinforces the methane signal in the first and third channels, while the middle channel (S) helps suppress false positives. Not only does it satisfy deep learning input requirements but also provides a signal-enhanced representation that enables models to better learn the complex spatial and spectral patterns of methane plumes, resulting in more reliable detection than using raw bands or a single-ratio image alone.

\subsection{Dataset Construction }

\begin{figure}[!tb]
    \centering
    \includegraphics[width=1\linewidth]{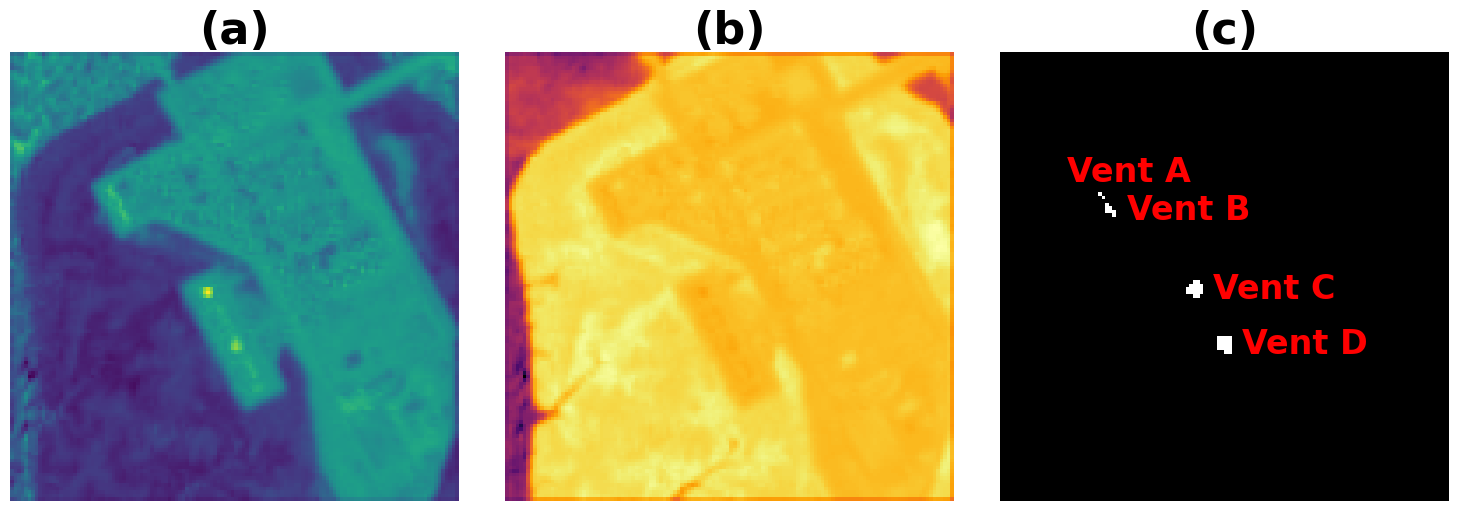}
    \vspace{-0.2cm}
    \caption{Example images from the study site on 23 July 2023, showing (a) the Varon ratio, (b) the Sanchez ratio, and (c) ground-truth annotations of methane plumes (labeled as Vent A–D) highlighting emission sources.}
     \vspace{-0.2cm}
    \label{fig:plume_gt}
\end{figure}

\textbf{Ground-Truth Generation:} Ground-truth annotations for methane plumes were developed through a systematic, semi-automated labelling process tailored to emission events at an Australian site in 2023  (see Fig.~\ref{fig:plume_gt}), ensuring precise and consistent spatial annotations for supervised learning. Initial plume boundaries were established using specialised contour detection functions and predefined concentration thresholds specific to the site's primary emission sources (Vent A-D). These automated contours were then meticulously refined through manual validation to capture distinct emission patterns and address complex scenarios, such as spatially overlapping plumes, while preserving individual source information. Validation against temporal and spatial emission records minimised subjectivity and ensured consistency between spectral signatures and documented events, establishing a reliable dataset for robust model training and evaluation.

\textbf{Background Data Generation:} In addition to plume-containing images, we curated methane-free background scenes from Sentinel-2 observations of regions like the ground-truth facility and other oil/gas sites during periods of no known emissions. These help the model recognise ``\textit{normal}" conditions and reduce false detections. No-plume datasets are generated using a regression-based simulation to characterise natural spectral behaviour under methane-free conditions. An adaptive percentile-based thresholding mechanism distinguishes methane-sensitive spectral regions from background areas, establishing robust SWIR band correlations \cite{pandey2023daily}. A linear regression model, trained on background pixels, predicts clean SWIR band 12 responses from SWIR band 11 inputs, preserving natural variability. Synthetic background data is enhanced with controlled Gaussian noise and cluster-based artefacts to simulate realistic observational and sensor noise conditions.

\textbf{Data Augmentation and Preprocessing:} To improve model generalisation, we applied a standard data augmentation pipeline, including geometric transformations (flips, rotations) and photometric adjustments (brightness, contrast). All input features were z-score normalised to ensure cross-scene consistency.

\subsection{Deep Learning Architectures and Training}
We evaluated three semantic segmentation architectures: U-Net \cite{Ronneberger2015UNetCN} with a MobileNetV2 \cite{Sandler2018MobileNetV2IR} encoder, U-Net with a ResNet34 \cite{He2015DeepRL} encoder, and SegNeXt \cite{Guo2022SegNeXtRC}. The U-Net is a strong baseline for segmentation tasks, and the choice of MobileNetV2 and ResNet34 encoders allowed us to evaluate the trade-off between computational efficiency and feature extraction depth. SegNeXt, a more recent architecture, blends convolutions with transformer-based attention mechanisms to capture global context.

All models were implemented in PyTorch, leveraging the ``segmentation-models-pytorch" library \cite{Iakubovskii:2019}. The encoders were initialised with ImageNet-pretrained weights to leverage transfer learning \cite{Krizhevsky2012ImageNetCW}. We used the Adam optimiser \cite{Kingma2014AdamAM} with an exponential decay learning rate scheduler for all training runs. To handle the significant class imbalance between small plume pixels and the large background, we employed a combined loss function of Focal Loss \cite{Lin2017FocalLF} and Dice Loss \cite{Milletar2016VNetFC} for the U-Net-ResNet34 and SegNeXt models. Hyperparameters for the loss function were optimised using a grid search \cite{Bergstra2012RandomSF}.

\begin{table}[!tb]
\centering
\caption{Model Performance on the Validation Set.}
\label{tab:results}
\begin{tabular}{@{}lcc@{}}
\toprule
\textbf{Model} & \textbf{F1-Score} & \textbf{IoU} \\
\midrule
U-Net + MobileNetV2 & 0.7720 & \textbf{0.7095} \\
U-Net + ResNet34 & \textbf{0.7839} & 0.6503 \\
SegNeXt & 0.7583 & 0.6188 \\
\bottomrule
\end{tabular}
\end{table}

\section{Experiments and Results}
%We assessed the performance of the models using the Dice coefficient (F1-score) and Intersection over Union (IoU), widely recognised metrics for segmentation tasks due to their robustness to class imbalance \cite{Mller2022TowardsAG}.

\subsection{Performance Metrics}
The efficacy of each network architecture was evaluated using the Dice coefficient (F1-score) and IoU \cite{xu2025infrared}, with the loss function guiding the learning process at each step. These metrics were selected for their ability to measure overlap between predicted segmentations and ground truth, while penalising false positives—a crucial consideration for class-imbalanced datasets such as methane plume imagery \cite{Mller2022TowardsAG}. In the context of methane plume detection, accurate segmentation of small, scattered plume regions demands precise boundary delineation and minimal background noise interference \cite{Chen2025MPSUNetAD, xu2025infrared}. The chosen metrics effectively capture these aspects, as higher Dice and IoU scores reflect improved edge accuracy and reduced false alarms in challenging segmentation scenarios \cite{xu2025infrared}.

\begin{figure}[!tb]
    \centering
    \includegraphics[width=0.9\linewidth]{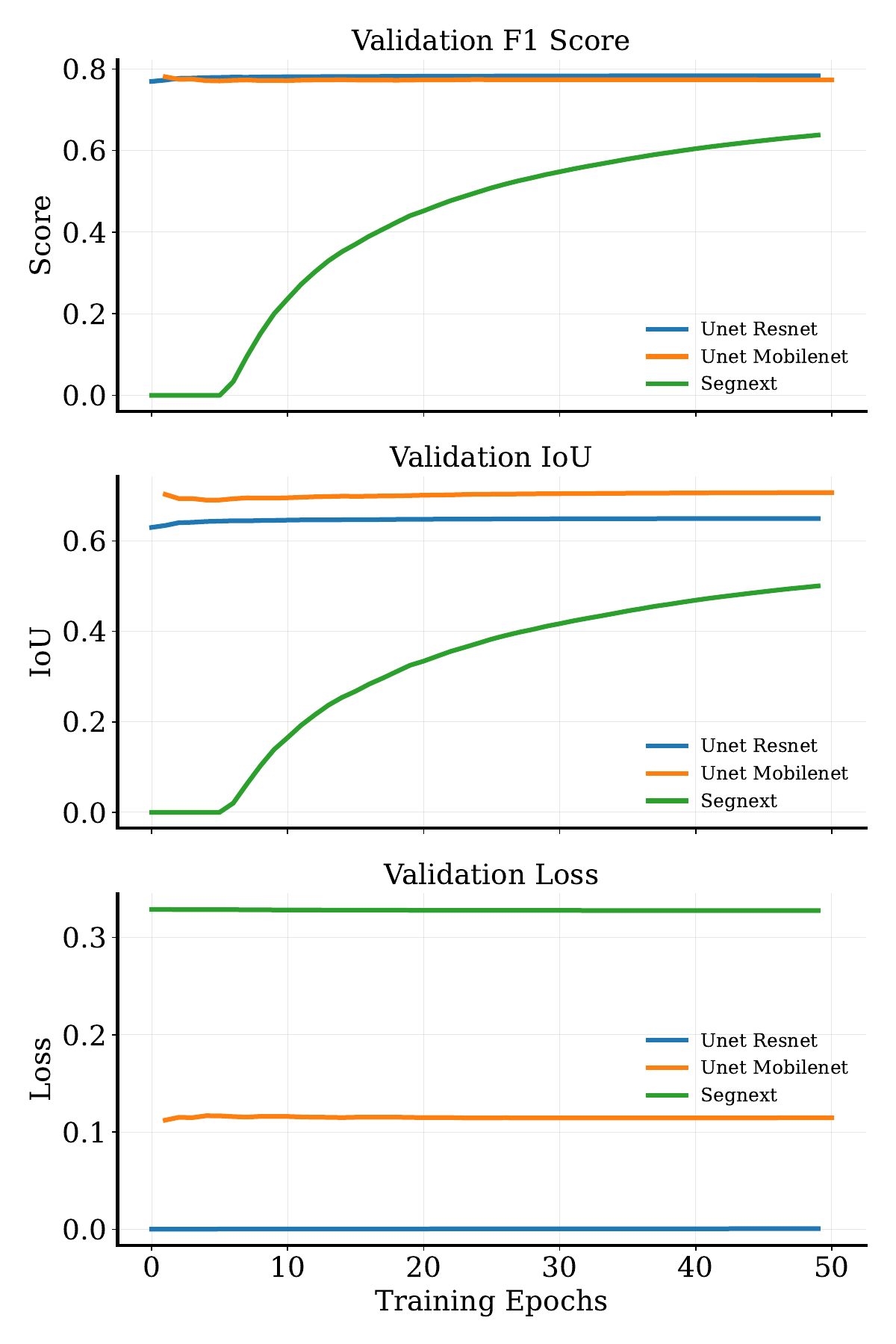}
    \vspace{-0.2cm}
    \caption{Learning curves for the three architectures, showing F1-score, IoU, and loss for validation sets. 
    The U-Net with ResNet34 encoder demonstrates the best and most stable performance.}
    \vspace{-0.2cm}
    \label{fig:learning_curves}
\end{figure}

\begin{figure*}[!tb]
  \centering
  \begin{subfigure}[b]{0.97\textwidth}
    \includegraphics[width=0.97\textwidth]{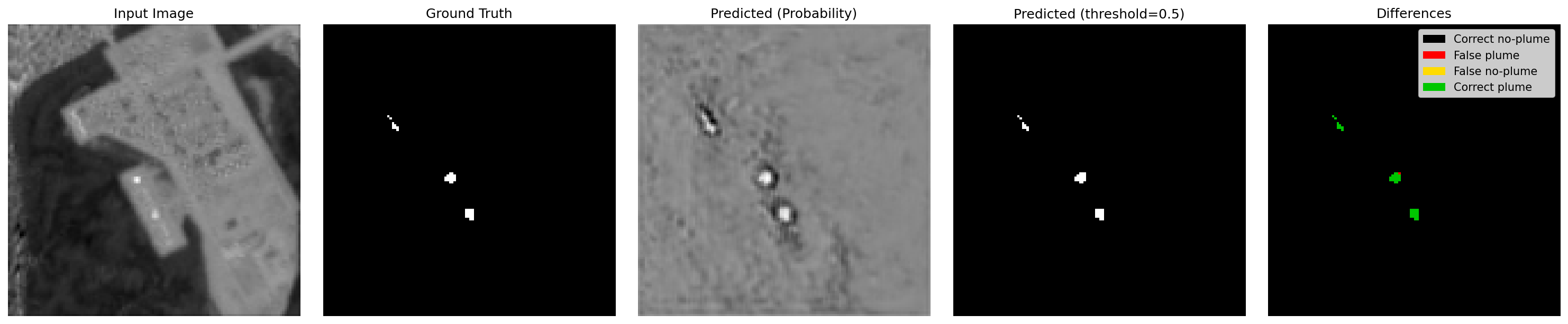}
    \vspace{-0.1cm}
    \caption{U-Net + MobileNetV2.}
  \end{subfigure}\par\vspace{2pt}
  \begin{subfigure}[b]{0.97\textwidth}
    \includegraphics[width=0.97\textwidth]{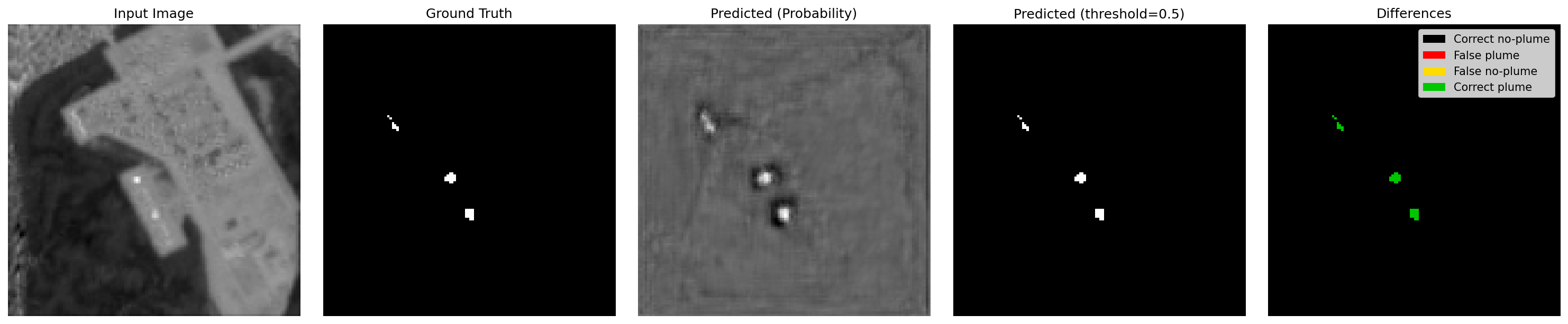}
    \vspace{-0.1cm}
    \caption{U-Net + ResNet34.}
  \end{subfigure}\par\vspace{2pt}
  \begin{subfigure}[b]{0.97\textwidth}
    \includegraphics[width=0.97\textwidth]{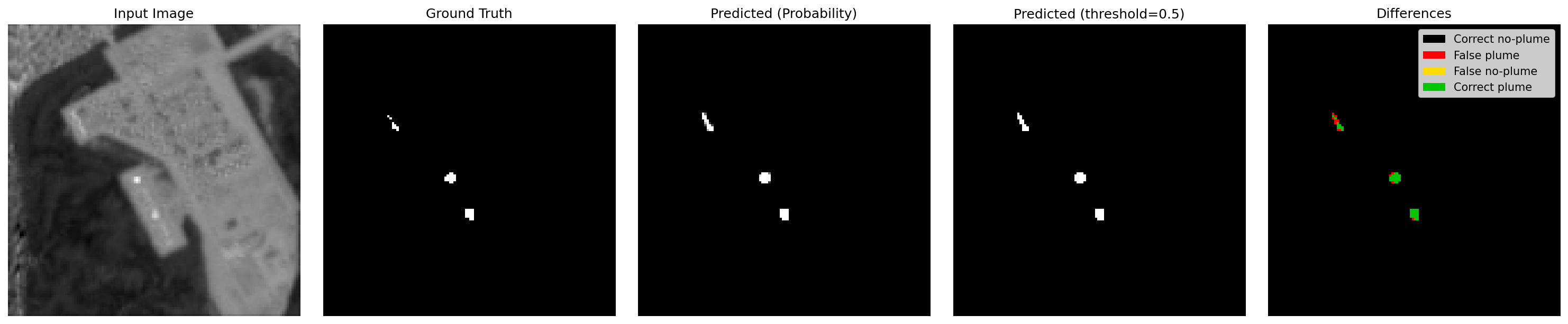}
    \vspace{-0.1cm}
    \caption{SegNeXt.}
  \end{subfigure}
  \vspace{-0.1cm}
  \caption{Qualitative comparison of model outputs. Each panel shows (from left to right): input, ground truth, probability map, binarised prediction, and a difference map (Green: correct plume, Red: false positive, Yellow: false negative). The ResNet34-powered U-Net (b) shows the most accurate segmentation.}
  \label{fig:qualitative_comparison}
\end{figure*}
\subsection{Quantitative Results}
We trained and evaluated the models on our curated dataset, summarising the validation set performance in Table~\ref{tab:results}. The learning curves for the three architectures are presented in Fig.~\ref{fig:learning_curves}.

The U-Net with MobileNetV2 (Fig.~\ref{fig:learning_curves}) exhibited notable improvements during training, achieving a high training F1-score of 0.9547 and training IoU of 0.9225. However, validation performance was moderate, with an F1-score of 0.7720 and an IoU of 0.7095. Although there was significant learning demonstrated, indicated by a consistent decrease in validation loss, the discrepancy between training and validation scores suggests potential overfitting and challenges in generalisation.

In contrast, the U-Net with the ResNet34 encoder (Fig.~\ref{fig:learning_curves}) demonstrated strong performance and stable learning curves, showcasing efficient learning without evident overfitting. It achieved a training F1-score of 0.9542 and a training IoU of 0.9131, closely matching MobileNetV2 in training performance but outperforming it in validation, attaining the highest validation F1-score of 0.7839. However, the validation IoU was comparatively lower at 0.6503, indicating a discrepancy between pixel-wise accuracy and overall detection capability. The residual connections within the ResNet34 encoder appear beneficial for capturing deep, relevant features effectively, enhancing segmentation robustness.

The SegNeXt model (Fig.~\ref{fig:learning_curves}) displayed steady learning progress, with training metrics at an F1-score of 0.7850 and IoU of 0.6492, and validation metrics of F1-score of 0.7583 and IoU of 0.6188. These results underscore the benefits of SegNeXt's transformer components in capturing and focusing on relevant features for plume segmentation, as noted by Guo et al.~\cite{Guo2022SegNeXtRC}, its validation performance did not surpass that of the other models. The lower scores suggest limitations in handling the complex segmentation challenges presented by the dataset.

Collectively, these results indicate distinct performance contrasts across the encoder architectures. MobileNetV2 demonstrated substantial training capabilities but faced generalisation issues. The ResNet34-based U-Net emerged as the strongest performer overall, particularly in validation F1-score. SegNeXt, while showing consistent learning behaviour and reasonable segmentation accuracy, remained behind the other two architectures. Therefore, the ResNet34 encoder stands out as the most robust and suitable model for methane plume detection in this study.

\subsection{Qualitative Results}
To further assess the performance of the trained models, we present a visual comparison of their segmentation capabilities on representative test scenes (Fig.~\ref{fig:qualitative_comparison}). These figures showcase input images, ground truth labels, predicted probability maps, binarised predictions, and difference visualisations. The ``Differences" panel employs a colour-coding scheme: black for accurate non-plume predictions, red for false positives (incorrectly classified plumes), yellow for false negatives (missed plumes), and green for correctly detected plume pixels.

The U-Net with MobileNetV2 encoder (Fig.~\ref{fig:qualitative_comparison},(a)) demonstrates effective segmentation performance, accurately detecting plume regions with no visible false negatives. However, a few scattered false positive indicate areas where the model misclassifies background regions as plume. These suggest MobileNetV2 effectively captures plume signals but has a mild tendency toward over-segmentation of background areas, which aligns with its quantitative metrics.

The ResNet34 U-Net model (Fig.~\ref{fig:qualitative_comparison},(b)) exhibits exceptional segmentation accuracy, precisely identifying plume regions with clearly defined boundaries. Significantly, it demonstrates neither false positives nor false negatives in the provided visualisation, underscoring the model’s remarkable precision and recall. These observations strongly corroborate its superior quantitative metrics, confirming ResNet34's capability in accurately differentiating plume signals from background structures without misclassifications.

The SegNeXt model (Fig.~\ref{fig:qualitative_comparison}\,(c)) detects main plume regions with few false negatives but slightly more scattered false positives than ResNet-34 U-Net. This is due to SegNeXt’s global attention mechanisms, which enable the model to consider and integrate information from all parts of the image when analyzing any specific region. While this global context helps the network capture extended or subtle plume structures, it can also amplify unrelated background patterns that happen to correlate across the scene—leading to those scattered false positives. The result is excellent plume coverage with a minor overestimation, matching its strong yet slightly lower quantitative scores.

\section{Conclusion}

Our experiments show that a U-Net architecture with a ResNet34 encoder provides the best balance of feature extraction and spatial localisation, achieving superior performance over more lightweight or complex models. Most importantly, we demonstrated that this approach can reliably detect methane plumes down to a single pixel, a significant step towards a fully automated and scalable monitoring system.

For future work, we consider three potential following directions: (i) extending the dataset and model testing to more diverse geographic regions and emission source types to improve generalizability; (ii) exploring model interpretability techniques (e.g., saliency maps) to build trust and better integrate these tools into operational workflows for climate scientists and regulators; and (iii) investigating multi-temporal or ensemble approaches that could leverage information across multiple satellite passes to further enhance detection sensitivity for the most intermittent and subtle emission events.

\section*{Acknowledgment}
A large language model (LLM) was utilised for checking grammar and English usage in this paper. 

\balance
\bibliography{ref.bib}
\end{document}